\documentclass[10pt,twocolumn,letterpaper]{article}

\usepackage{iccv}
\usepackage{times}
\usepackage{epsfig}
\usepackage{graphicx}
\usepackage{amsmath}
\usepackage{amssymb}

\usepackage{algorithm}
\usepackage{algorithmicx}
\usepackage{amsmath,amssymb}
\usepackage{algpseudocode}

\usepackage{subfig}
\usepackage[export]{adjustbox}
\usepackage{makecell}
\usepackage{booktabs} 
\usepackage{multirow}\usepackage{xcolor}
\usepackage{authblk}

\newcommand{\vct}[1]{\ensuremath{\boldsymbol{#1}}} 

\newcommand{\set}[1]{\ensuremath{\mathcal{#1}}}
\newcommand{\con}[1]{\ensuremath{\mathsf{#1}}}

\newcommand{\argmax}{\operatornamewithlimits{\arg\,\max}}

\newcommand{\SVM}{SVM\xspace}
\newcommand{\SVMRBF}{SVM-RBF\xspace}
\newcommand{\SVMADV}{SVM-adv\xspace}

\newcommand{\DIcub}{\mbox{\textsl{iCubWorld28}}\xspace}
\newcommand{\DIcubRED}{\mbox{\textsl{iCubWorld7}}\xspace}

\newcommand{\myparagraph}[1]{\smallskip \noindent \textbf{#1}}

\usepackage[pagebackref=true,breaklinks=true,letterpaper=true,colorlinks,bookmarks=false]{hyperref}

\iccvfinalcopy 

\def\iccvPaperID{****} 

\ificcvfinal\fi
\begin{document}

\title{Is Deep Learning Safe for Robot Vision?\\ Adversarial Examples against the iCub Humanoid}

\author[1]{Marco Melis}
\author[1]{Ambra Demontis}
\author[1,2]{Battista Biggio}
\author[3]{	Gavin Brown}
\author[1]{	\\Giorgio Fumera}
\author[1,2]{Fabio Roli}

\affil[1] {PRA Lab, Department of Electrical and Electronic Engineering, University of Cagliari, Italy, {\small \url{http://pralab.diee.unica.it/en}}}
\affil[2] {Pluribus One, Italy, {\small \url{http://www.pluribus-one.it}}}
\affil[3] {School of Computer Science, University of Manchester, UK}

\renewcommand\Authands{ and }

\maketitle

\begin{abstract}
Deep neural networks have been widely adopted in recent years, exhibiting impressive performances in several application domains.
It has however been shown that they can be fooled by adversarial examples, i.e., images altered by a barely-perceivable adversarial noise, carefully crafted to mislead classification. 
In this work, we aim to evaluate the extent to which robot-vision systems embodying deep-learning algorithms are vulnerable to adversarial examples, and propose a computationally efficient countermeasure to mitigate this threat, based on rejecting classification of anomalous inputs.
We then provide a clearer understanding of the safety properties of deep networks through an intuitive empirical analysis, showing that the mapping learned by such networks essentially violates the smoothness assumption of learning algorithms. We finally discuss the main limitations of this work, including the creation of real-world adversarial examples, and sketch promising research directions.\footnote{Accepted for publication at the ICCV 2017 Workshop on Vision in Practice on Autonomous Robots (ViPAR).}
\end{abstract}

\section{Introduction} 

After decades of research spent in exploring different approaches, ranging from search algorithms, expert and rule-based systems to more modern machine-learning algorithms, several problems involving the use of an artificial intelligence have been finally tackled through the introduction of a novel paradigm shift based on \emph{data-driven} artificial intelligence technologies. In fact, due to the increasing popularity and use of the modern Internet, along with the powerful computing resources available nowadays, it has been possible to extract meaningful knowledge from the huge amount of data collected online, from images to videos, text and speech data~\cite{cristianini2016intelligence}.
Deep learning algorithms have provided an important resource in this respect. Their flexibility to deal with different kinds of input data, along with their learning capacity, have made them a powerful instrument to successfully tackle challenging applications, reporting impressive performance on several tasks in computer vision, speech recognition and human-robot interactions~\cite{hinton2012deep,pasquale2015teaching}.

\begin{figure*}[t]
	\centering
	\includegraphics[width=0.99\textwidth]{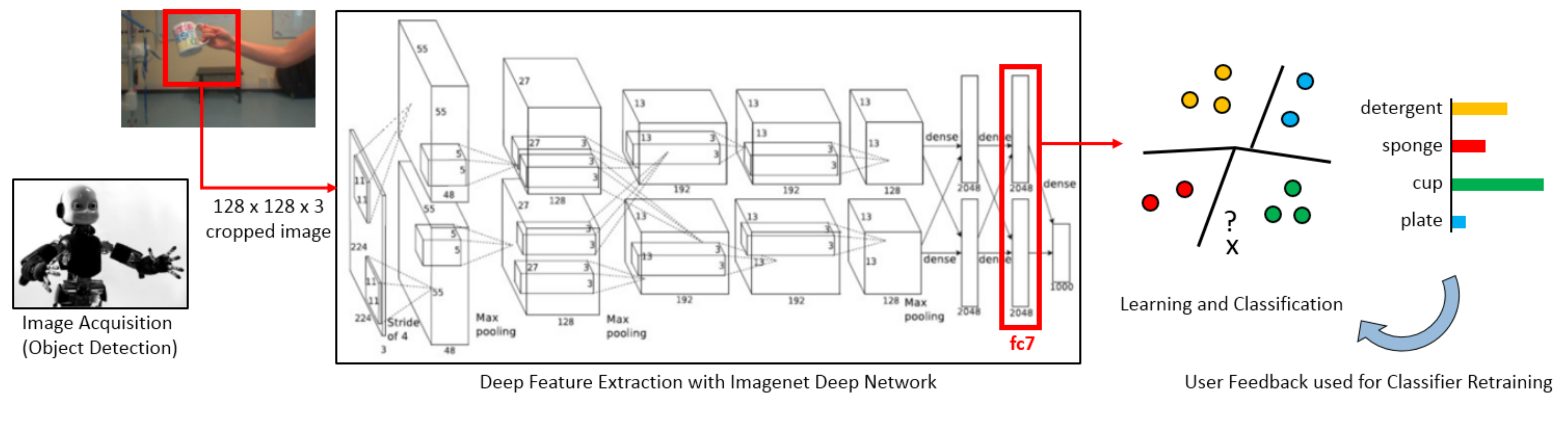}
\caption{Architecture of the iCub robot-vision system~\cite{pasquale2015teaching}. After image acquisition, a region of interest containing the object is cropped and processed by the ImageNet deep network~\cite{alex12-nips}.
	The deep features extracted from the penultimate layer of such network (fc7) are then used as input to the classification algorithm to perform the recognition task, in which the probabilities that the object belongs to each (known) class are reported.
	A human annotator can then validate or correct decisions, and the classification algorithm can be updated accordingly; for instance, to learn that an object belongs to a never-before-seen class.}
	\label{fig:iCub}
\end{figure*}

Despite their undiscussed success in several real-world applications, several open problems remain to be addressed.
Research work has been investigating how to interpret decisions taken by deep learning algorithms, unveiling the patterns learned by deep networks at each layer~\cite{zeiler2014visualizing,mahendran2015understanding}. Although a significant progress have been made in this direction, and it is now clear that such networks gradually learn more abstract concepts (\eg, from detecting elementary shapes in images to more abstract notions of objects or animals), a relevant effort is still required to gain deeper insights.
This is also important to understand why such algorithms may be \emph{vulnerable} to the presence of \emph{adversarial examples}, \ie, input data that are slightly modified to mislead classification by the addition of an almost-imperceptible adversarial noise~\cite{luo2015foveation,moosavi2016deepfool}. 
The presence of adversarial examples have been shown on a variety of tasks, including object recognition in images, handwritten digit recognition, and face recognition~\cite{sharif2016accessorize,szegedy14-iclr,goodfellow15-iclr,moosavi2016deepfool,papernot16-sp}.

In this work, we are the first to show that robot-vision systems based on deep learning algorithms are also vulnerable to this potential threat. 
This is a crucial problem, as embodied agents have a more direct, physical interaction with humans than virtual agents, and the damage caused by adversarial examples in this context can thus be much more concerning.
To demonstrate this vulnerability, we focus on a case study involving the iCub humanoid robot (Sect.~\ref{sec:icub})~\cite{metta2008icub,pasquale2015teaching}. A peculiarity of humanoid robots is that they have to be able to learn in an online fashion, from the stimuli received during their exploration of the surrounding environment. For this reason, a crucial requirement for them is to embody completely the acquired knowledge, and a reliable and efficient learning paradigm.
As discussed in previous work~\cite{pasquale2015teaching},
this is a conflicting goal with the current state of deep learning algorithms, which are too computationally and power demanding to be fully embodied by a humanoid robot.
For this reason, the authors in~\cite{pasquale2015teaching} have proposed to use a pre-trained deep network for object recognition to perform feature extraction (essentially considering as the feature vector for the detected object one of the last convolutional layers in the deep network), and then train a multiclass classifier on such feature representation.

The first contribution of this work is to show the vulnerability of these kinds of robot-vision system to adversarial examples. To this end, we propose an alternative algorithm for the generation of adversarial examples (Sect.~\ref{sec:evasion}), which extends previous work on the evasion of binary to multiclass classifiers~\cite{biggio13-ecml}. Conversely to previous work dealing with the generation of adversarial examples based on minimum-distance perturbations~\cite{szegedy14-iclr,goodfellow15-iclr,moosavi2016deepfool,papernot16-sp}, our algorithm enables creating adversarial examples misclassified with higher confidence, under a maximum input perturbation, for which devising proper countermeasures is also more difficult. This allows one to assess classifier security more thoroughly, by evaluating the probability of evading detection as a function of the maximum input perturbation.
Notably, it also allows manipulating only a region of interest in the input image, such that creating real-world adversarial examples becomes easier; \eg, one may only modify some image pixels corresponding to a sticker that can be subsequently applied to the object of interest.

The second contribution of this work is the proposal of a computationally-efficient countermeasure, inspired from work on classification with the reject option and open-set recognition, to mitigate the threat posed by adversarial examples.
Its underlying idea is to detect and reject the so-called blind-spot evasion points, \ie, samples which are sufficiently far from known training data. This countermeasure is particularly suited to our case study, as it  requires modifying only the learning algorithm applied on top of the deep feature representation, \ie, only the output layer (Sect.~\ref{sec:improving_icub}).

We then report an empirical evaluation (Sect.~\ref{sec:exp}) showing that the iCub humanoid is vulnerable to adversarial examples and to which extent our proposed countermeasure can improve its security.
In particular, although it does not completely address the vulnerability of such system to adversarial examples, it requires one to significantly increase the amount of perturbation on the input images to reach a comparable probability of misleading a correct object recognition.
To better understand the reason behind this phenomenon, we provide a further, simple and intuitive empirical analysis, showing that the mapping learned by the deep network used for deep feature extraction essentially violates the smoothness assumption of learning techniques in the input space. 
This means that, in practice, for a sufficiently high amount of perturbation, the proposed algorithm creates adversarial examples that are mapped onto a region of the deep feature space which is densely populated by training examples of a different class.
Accordingly, only modifying the classification algorithm on top of the pre-trained deep features (without re-training the underlying deep network) may not be sufficient in this case. 

We conclude this paper discussing related work (Sect.~\ref{sec:related_work}), and relevant future research directions (Sect.~\ref{sec:conclusion}).

\section{The iCub Humanoid} 
\label{sec:icub}

Our case study focuses on the iCub humanoid, as it provides a cognitive humanoid robotic platform well suited to our task~\cite{metta2008icub,pasquale2015teaching}.
In particular, the visual recognition system of this humanoid relies on deep learning technologies to interact with the surrounding environment, enabling it to detect and to recognize \emph{known} objects, \ie, objects that have been verbally annotated in a previous session by a human teacher.
Furthermore, iCub is capable of performing \emph{online} learning, \ie, after classification, it asks to the human teacher whether the corresponding decision is correct. If the decision is wrong (\eg, in the case of an object belonging to a never-before-seen class), the human teacher can provide feedback to the robot, which in turn updates its classification model through online or incremental learning techniques (\eg, by expanding the set of known object classes).
This a clear example of how a robot can learn from experience to improve its capabilities, \ie, a key aspect of why embodying knowledge within robots is of crucial relevance for these tasks~\cite{pasquale2015teaching}.
However, given the limited hardware and power resources of the humanoid, it is clear that retraining the whole deep learning infrastructure becomes too computationally demanding.
For this reason, the visual system of iCub exploits the pre-trained ImageNet deep network~\cite{alex12-nips} 
only for extracting a set of deep features (from one of the highest convolutional layers) and uses this feature vector to represent the object detected by iCub in the input image.
As described in Fig.~\ref{fig:iCub}, this deep feature vector is then classified using a separate classifier, which can be retrained online in an efficient manner when feedback from the human annotator is received.
In particular, in~\cite{pasquale2015teaching} this classifier is implemented using a one-versus-all scheme to combine a set of $\con c$ linear classifiers, being $\con c$ the number of known classes.
Let us denote the pixel values of the input image (in raster-scan order) with $\vct x \in \set X \subseteq \mathbb R^{\con d}$ (where $\con d = 128 \times 128 \times 3$), and the discriminant functions of the aforementioned one-versus-all linear classifiers as $f_{1}(\vct x), \ldots, f_{\con c}(\vct x)$. Accordingly, the predicted class $c^{\star}$ is determined as the class whose discriminant function for that sample is maximum:
\begin{align}
c^{\star} = \argmax_{k=1, \ldots, \con c}  f_k(\vct x ) \, .
\label{eq:ev2}
\end{align}
The linear classifiers used for this purpose include Support Vector Machines (SVMs) and Recursive Least Square (RLS) classifiers, as both can be efficiently updated online~\cite{pasquale2015teaching}.
Notably, previous work has shown that replacing the softmax layer in deep networks with a multiclass SVM can be effective also in different applications~\cite{tang13-deepsvm}.

\section{Adversarial Security Evaluation}
\label{sec:evasion}

We discuss here our proposal to assess the security of robot-vision systems to adversarial examples.
As in previous work addressing the issue of evaluating security of machine-learning algorithms~\cite{barreno10,huang11,biggio14-tkde,biggio13-ecml,srndic14}, our underlying idea is to evaluate the maximum recognition accuracy degradation against an increasing maximum admissible level of perturbation of the input images.
This is rather different than previous work in which adversarial examples correspond to minimally-perturbed samples that are wrongly classified~\cite{szegedy14-iclr,goodfellow15-iclr,moosavi2016deepfool,papernot16-sp}. As we will see in our experiments,
besides providing a more complete evaluation of system security against adversarial examples,
our attack strategy also highlights additional interesting insights on system security, including the identification of vulnerabilities in the feature representation (rather than in the classification algorithm itself) through the creation of adversarial examples that are \emph{indistinguishable} from training samples of a different class.

Our approach is based on extending the work in~\cite{biggio13-ecml} for evasion of binary classifiers to the multiclass case. To this end, we define two possible evasion settings, \ie, ways of creating adversarial examples, which further differentiate our technique from previous work on the creation of minimally-perturbed adversarial examples~\cite{szegedy14-iclr,goodfellow15-iclr,moosavi2016deepfool,papernot16-sp}.
In particular, we consider an \emph{error-generic} and an \emph{error-specific} evasion setting.
In the \emph{error-generic} scenario, the attacker is interested in misleading classification, regardless of the output class predicted by the classifier for the adversarial examples; \eg, for a known terrorist the goal may be to evade detection by a video surveillance system, regardless of the identity that may be erroneously associated to his/her face.
Conversely, in the \emph{error-specific} setting, the attacker still aims to mislead classification, but requiring the adversarial examples to be misclassified as a specific, target class; \eg, imagine an attacker aiming to impersonate a specific user.\footnote{In~\cite{papernot16-sp}, the authors defined \emph{targeted} and \emph{indiscriminate} attacks depending on whether the attacker aims to cause \emph{specific} or \emph{generic} errors, similarly to our settings. Here we do not follow their naming convention, as it causes confusion with the interpretation of \emph{targeted} and \emph{indiscriminate} attacks introduced in previous work~\cite{barreno10,huang11,biggio14-tkde}.}

The two settings can be formalized in terms of two distinct optimization problems, though using the same formulation for the objective function $\Omega(\vct x)$:
\begin{equation}
\Omega(\vct x) = f_{k}(\vct x) - \max_{l \neq k} f_{l}(\vct x) \, .
\label{eq:omega}
\end{equation}
This function essentially represents a difference between a preselected discriminant function (associated to class $k$) and the competing one, \ie, the one exhibiting the highest value at $\vct x$ among the remaining $\con c -1$ classes (\ie, all classes $\{1, \ldots, \con c\}$ except $k$). We discuss below how class $k$ is chosen in the two considered settings.

\myparagraph{Error-generic Evasion.} In this case, the optimization problem can be formulated as:
\begin{eqnarray}
\label{eq:indiscriminate-1}
\min_{\vct x^{\prime}}  & & \Omega(\vct x^{\prime}) \, ,\\
\label{eq:indiscriminate-2}
{\rm s.t. } && d(\vct x, \vct x^{\prime}) \leq d_{\rm max} \, ,\\
\label{eq:indiscriminate-3}
&& \vct x_{\rm lb} \preceq \vct x^{\prime} \preceq \vct x_{\rm ub} \, ,
\end{eqnarray}
where $f_{k}(\vct x)$ in the objective function $\Omega(\vct x)$ (Eq.~\ref{eq:omega}) denotes the discriminant function associated to the true class of the source sample $\vct x$, and
$d(\vct x, \vct x^{\prime}) \leq d_{\rm max}$ represents a constraint on the maximum input perturbation $d_{\rm max}$ between $\vct x$ (\ie, the input image) and the corresponding modified adversarial example $\vct x^{\prime}$, given in terms of a distance in the input space. Normally, the $\ell_{2}$ distance between pixel values is used as the function $d(\cdot, \cdot)$, but other metrics can be also adopted (\eg, one may use an $\ell_{1}$-based constraint to inject a sparse adversarial noise rather than a slight image blurring as that caused by the $\ell_{2}$-based constraint)~\cite{demontis16-spr,russu16-aisec}. 
The box constraint $\vct x_{\rm lb} \preceq \vct x^{\prime} \preceq \vct x_{\rm ub}$ (where $\vct u \preceq \vct v$ means that each element of $\vct u$ has to be not greater than the corresponding element in $\vct v$) is optional, and can be used to bound the input values $\vct x$ of the adversarial examples; \eg, each pixel value in images is bounded between $0$ and $255$.
Nevertheless, the box constraint can be also used to manipulate only some pixels in the image.
For example, if some pixels should not be manipulated, one can set the corresponding values of $\vct x_{\rm lb}$ and $\vct x_{\rm ub}$ equal to those of $\vct x$.
 This is of crucial importance for creating real-world adversarial examples, as it allows one to avoid manipulating pixels which do not belong to the object of interest. For instance, this may enable one to create an ``unusual'' sticker to be attached to an \emph{adversarial} object, similarly to the idea exploited in~\cite{sharif2016accessorize} for the creation of wearable objects used to fool face recognition systems.

\begin{figure}[t]
\centering
\includegraphics[width=0.43\columnwidth]{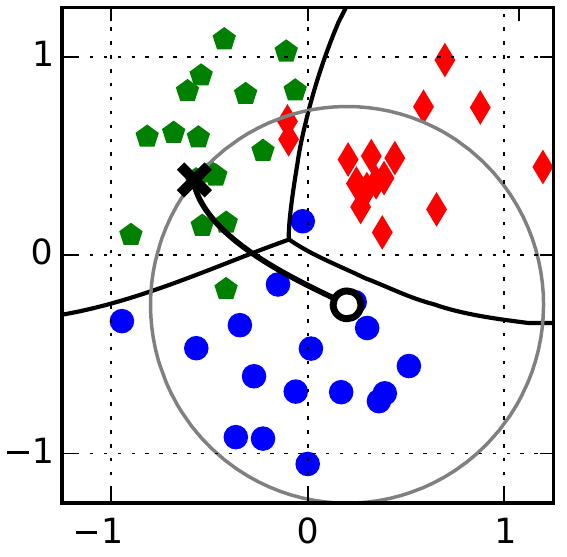}
\includegraphics[width=0.43\columnwidth]{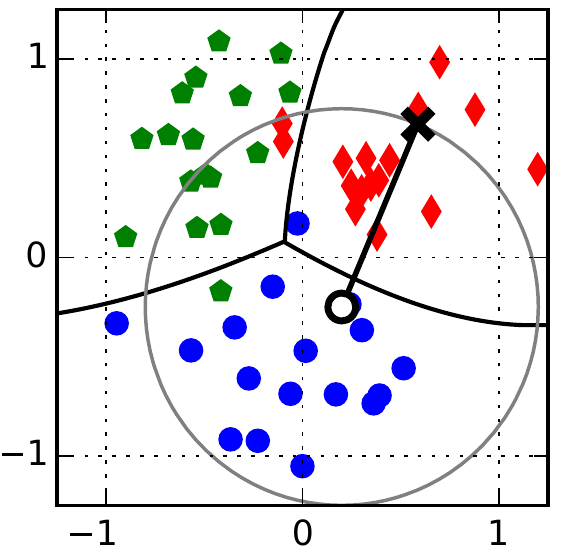}
\caption{Error-specific (\emph{left}) and error-generic (\emph{right}) evasion of a multiclass SVM with the Radial Basis Function (RBF) kernel.
Decision boundaries among the three classes (blue, red and green points) are shown as black solid lines. In the error-specific case, the initial (blue) sample is shifted towards the green class (selected as the target one).
In the error-generic case, instead, it is shifted towards the red class, as it is the closest class to the initial sample. The $\ell_{2}$ distance constraint is also shown as a gray circle.}
\label{fig:eva-targeted-indiscriminate}
\end{figure}

\myparagraph{Error-specific Evasion.} The problem of error-specific evasion is formulated as the error-generic evasion problem in Eqs.~\eqref{eq:indiscriminate-1}-\eqref{eq:indiscriminate-3}, with the only differences that: ($i$) the objective function is \emph{maximized}; and ($ii$) $f_{k}$ denotes the discriminant function associated to the targeted class, \ie, the class which the adversarial example should be assigned to.

An example of the different behavior exhibited by the two attacks is given in Fig.~\ref{fig:eva-targeted-indiscriminate}.
Both attacks are constructed using the simple gradient-based algorithm given as Algorithm~\ref{alg:evasion}.
The basic idea is to update the adversarial example by following the steepest descent (or ascent) direction (depending on whether we are considering error-generic or error-specific evasion), and use a projection operator $\Pi$ to keep the updated point within the feasible domain (given by the intersection of the box and the $\ell_{2}$ constraint).

\myparagraph{Gradient computation.} One key issue of the aforementioned algorithm is the computation of the gradient of $\Omega(\vct x)$, which involve the gradients of the discriminant function $f_{i}(\vct x)$ for $i \in {1, \ldots, \con c}$. It is not difficult to see that this can be computed using the chain rule to decouple the gradient of the discriminant function of the classifier trained on the deep feature space and the gradient of the deep network used for feature extraction, as 
$\nabla f_{i}(\vct x) = \frac{ \partial f_{i}(\vct z)}{ \partial \vct z }  \frac{ \partial \vct z}{ \partial \vct x }$, being $\vct z \in \mathbb R^{\con m}$ the set of deep features. In our case study, these are the $\con m = 4,096$ values extracted from layer fc7 (see Fig.~\ref{fig:iCub}).
Notably, the gradient of the deep network $\frac{ \partial \vct z}{ \partial \vct x}$ is readily available through automatic differentiation, as also highlighted in previous work~\cite{szegedy14-iclr,goodfellow15-iclr,moosavi2016deepfool,papernot16-sp}, whereas the availability of the gradient $\frac{ \partial f_{i}(\vct z)}{ \partial \vct z }$ depends on whether the chosen classifier is differentiable or not.
Several of the most used classifiers are differentiable, including, \eg, SVMs with differentiable kernels (we refer the reader to~\cite{biggio13-ecml} for further details). Nevertheless, if the classifier is not differentiable (\eg, like in the case of decision trees), one may use a surrogate differentiable classifier to approximate it, as also suggested in~\cite{biggio13-ecml,demontis16-spr,russu16-aisec}.

\begin{algorithm}[t]
	\caption{Computation of Adversarial Examples}
	\label{alg:evasion}
	\begin{algorithmic}[1]
		\Require $\vct x_{0}$: the input image;  $\eta$: the step size; $r \in \{-1,+1\}$: variable set to $-1$ ($+1$) for error-generic (error-specific) evasion; $\epsilon > 0$: a small number. 
		\Ensure $\vct x^{\prime}$: the adversarial example.
		\State $\vct x^{\prime} \gets \vct x_{0}$
		\Repeat
		\State $\vct x \gets \vct x^{\prime}$, and  $ \vct x^\prime \gets  \Pi \left ( \vct x +  r \eta \nabla \Omega(\vct x) \right ) $
		\Until{$ | \Omega (\vct x^\prime) - \Omega (\vct x) | \le \epsilon $}
		\State \Return $\vct x^\prime$
	\end{algorithmic}
\end{algorithm}

\section{Classifier Security to Adversarial Examples}
\label{sec:improving_icub}

\begin{figure*}[t]
\centering
\includegraphics[width=0.23\textwidth]{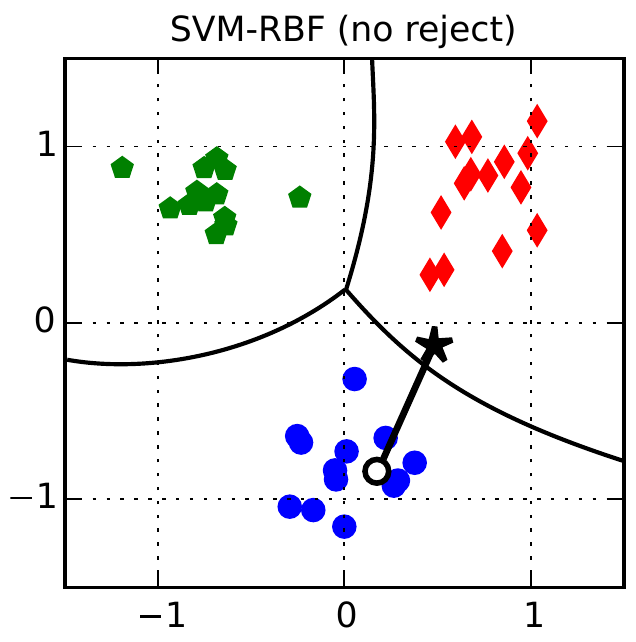}
\includegraphics[width=0.23\textwidth]{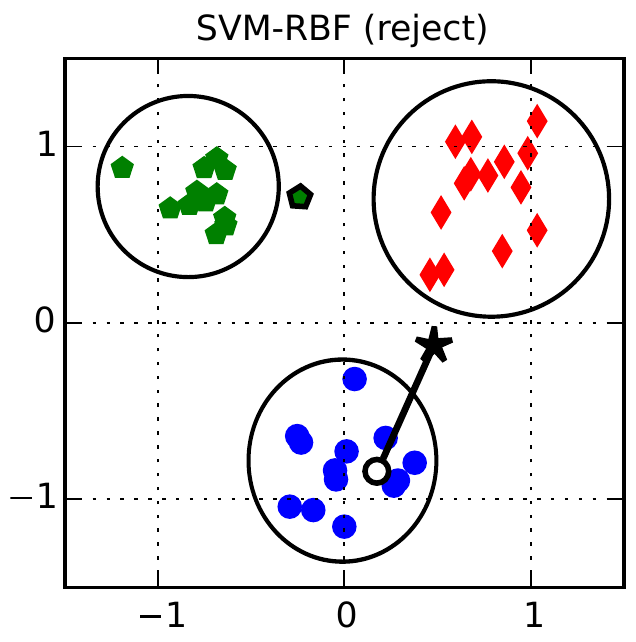}
\includegraphics[width=0.23\textwidth]{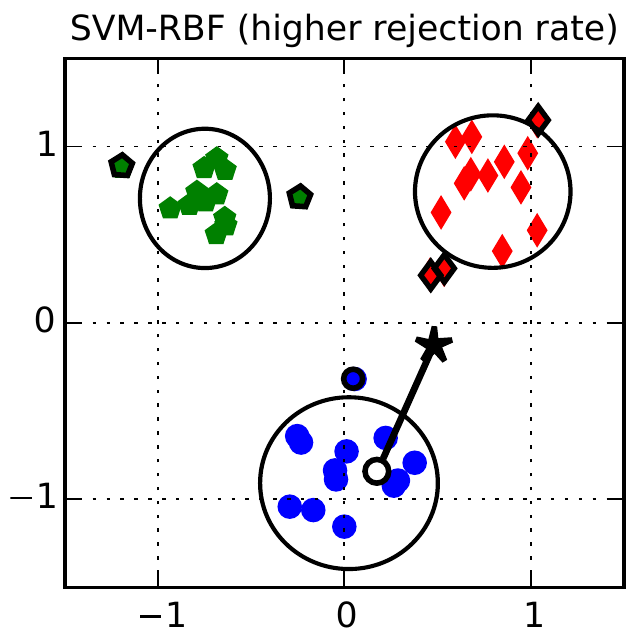}
\caption{Conceptual representation of our idea behind improving iCub security to adversarial examples, using multiclass SVMs with RBF kernels (\SVMRBF), without reject option (no defense, \emph{left}), with reject option (\emph{middle}), and with modified thresholds to increase the rejection rate (\emph{right}). Rejected samples are highlighted with black contours. The adversarial example (black star) is misclassified as a red sample by \SVMRBF (left plot), while \SVMRBF with reject option correctly identifies it as an adversarial example (middle plot). Rejection thresholds can be modified to increase classifier security (right plot), though at the expense of misclassifying more legitimate (\ie, non-manipulated) samples.}\label{fig:svm-rbf-reject}
\end{figure*}

If the evasion algorithm drives the adversarial examples deeply into regions populated by known training classes (as shown in Fig.~\ref{fig:eva-targeted-indiscriminate}), there is no much one can do to correctly identify them from the rest of the data by only re-training or modifying the classifier, \ie, modifying the shape of the decision boundaries in the feature space. We propose to consider this problem as an intrinsic \emph{vulnerability} of the \emph{feature representation}: if the feature vector of an adversarial example becomes \emph{indistinguishable} from those of the training samples of a different class, it can only be detected by using a different feature representation (\ie, in the case of iCub, this would require at least re-training the underlying deep network responsible for deep feature extraction).\footnote{Here, we only refer to the classifier trained on top of the deep feature representation as the \emph{classification algorithm}. This definition excludes the pre-trained deep network used for feature extraction in iCub, as it is not re-trained online.}
However, this is not always the case, especially in high-dimensional spaces, or if classes are separated with a sufficiently high \emph{margin}. In this case, as depicted in Fig.~\ref{fig:svm-rbf-reject}, there may be very large regions of the feature space which are only scarcely populated by data, although being associated (potentially also with high confidence) to known classes by the learning algorithm. Accordingly, adversarial examples may quite reasonably end up in such regions while also successfully fooling detection. These samples are often referred to as \emph{blind-spot} evasion samples, as they are capable of misleading classification, but in regions of the space which are far from the rest of the training data~\cite{huang11,szegedy14-iclr}. Conversely to the case of \emph{indistinguishable} adversarial examples, \emph{blind-spot} adversarial examples can be detected by only modifying the classifier (\ie, without re-training the underlying deep network used by iCub).
Accordingly, we propose to consider the problem of blind-spot adversarial examples as an intrinsic vulnerability of the classification algorithm. Different approaches have been proposed based on modifying the classifier, ranging from 1.5-class classification (based on the combination of anomaly detectors and two-class classifiers)~\cite{biggio15-mcs} to open-set recognition techniques~\cite{scheirer14,bendale2016towards}.

We propose here a more direct approach, based on the same idea underlying the notion of classification with a reject option, and leveraging some concepts from open-set recognition. In particular, we consider SVMs with RBF kernels to implement the multiclass classifier in our case study, as these SVMs belong to the so-called class of Compact Abating Probability (CAP) models~\cite{scheirer14} (\ie, classifiers whose discriminant function decreases while getting farther from the training data).
Then, by applying a simple rejection mechanism on their discriminant function, we can identify samples which are far enough from the rest of the training data, \ie, blind-spot adversarial examples.
Our idea is thus to modify the decision rule in Eq.~\eqref{eq:ev2} as:
\begin{equation}
c^{\star} = \argmax_{k=1, \ldots, \con c}  f_k(\vct x ) \, , \, {\rm only } \, {\rm if } \, f_{c^{\star}}(\vct x) > 0 \, ,
\label{eq:ev-rej}
\end{equation}
otherwise classify $\vct x$ as an adversarial example (\ie, a novel class).
In practice this means that, if no classifier assigns the sample to an existing class (\ie, no value of $f$ is positive),
then we simply categorize it as an adversarial example. In our specific case study, iCub may reject classification and ask the human annotator to label the example correctly.
Notably, the threshold of each discriminant function (\ie, the biases of the one-versus-all SVMs) can be adjusted to tune the trade-off between the rejection rate of adversarial examples and the fraction of incorrectly-rejected samples (which are not adversarially manipulated), as shown in Fig.~\ref{fig:svm-rbf-reject}.

\section{Experimental Analysis} 
\label{sec:exp}
In this section we report the results of the security evaluation performed on the iCub system (see Sect. \ref{sec:icub}) along with few adversarial examples to show how the proposed evasion algorithm can be exploited to create real-world attack samples. We then provide a conceptual representation and an empirical analysis to explain why neural networks are easily fooled and how our defense mechanism can improve their security in this context.

\begin{figure}
\centering
\begin{adjustbox}{width=\columnwidth}
\addtolength{\tabcolsep}{-5pt}
\begin{tabular}{ccccccc}
\includegraphics[width=.75in]{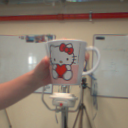} &
\includegraphics[width=.75in]{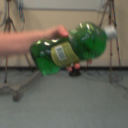} &
\includegraphics[width=.75in]{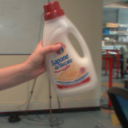} &
\includegraphics[width=.75in]{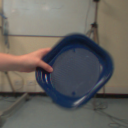} &
\includegraphics[width=.75in]{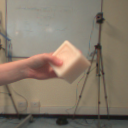} &
\includegraphics[width=.75in]{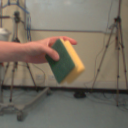} &
\includegraphics[width=.75in]{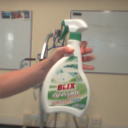} \\
\includegraphics[width=.75in]{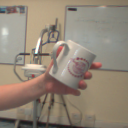} &
\includegraphics[width=.75in]{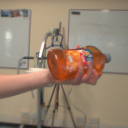} &
\includegraphics[width=.75in]{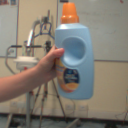} &
\includegraphics[width=.75in]{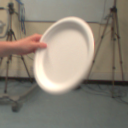} &
\includegraphics[width=.75in]{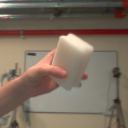} &
\includegraphics[width=.75in]{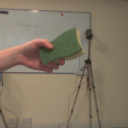} &
\includegraphics[width=.75in]{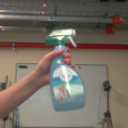} \\
\includegraphics[width=.75in,cfbox=red 2pt -1pt]{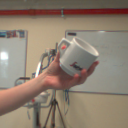} &
\includegraphics[width=.75in,cfbox=red 2pt -1pt]{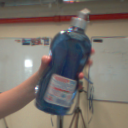} &
\includegraphics[width=.75in,cfbox=red 2pt -1pt]{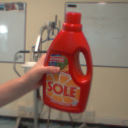} &
\includegraphics[width=.75in,cfbox=red 2pt -1pt]{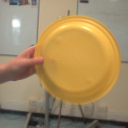} &
\includegraphics[width=.75in,cfbox=red 2pt -1pt]{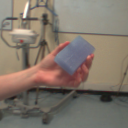} &
\includegraphics[width=.75in,cfbox=red 2pt -1pt]{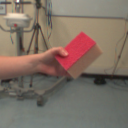} &
\includegraphics[width=.75in,cfbox=red 2pt -1pt]{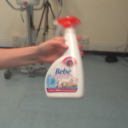} \\
\includegraphics[width=.75in]{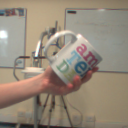} &
\includegraphics[width=.75in]{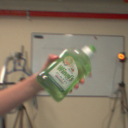} &
\includegraphics[width=.75in]{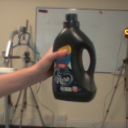} &
\includegraphics[width=.75in]{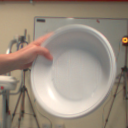} &
\includegraphics[width=.75in]{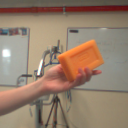} &
\includegraphics[width=.75in]{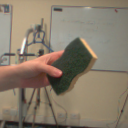} &
\includegraphics[width=.75in]{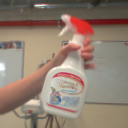} \\
Cup & \makecell{Dishwashing\\Detergent} & \makecell{Laundry\\Detergent} & Plate & Soap & Sponge & Sprayer
\addtolength{\tabcolsep}{5pt}
\end{tabular}
\end{adjustbox}
\caption{Example images (one per class) from the \DIcub dataset, and subset of classes used in the \DIcubRED dataset (highlighted in red).}\label{fig:icub-ds}
\end{figure}

\myparagraph{Experimental Setup.} Our analysis has been performed using the \DIcub dataset~\cite{pasquale2015teaching}, consisting of $28$ different classes which include $7$ different objects (cup, plate, \etc.) of $4$ different kinds each (\eg, cup1, cup2, \etc.), as shown in Fig.~\ref{fig:icub-ds}.
Each object was shown to iCub which automatically detected it and cropped the corresponding object image.
Four acquisition sessions were performed in four different days, ending up with approximately $20,000$ images for training and test sets.
As shown in~\cite{pasquale2015teaching}, it is very difficult for iCub to be able to distinguish such slight category distinctions, like different kinds of cups. For this reason, we also consider here a reduced dataset, \DIcubRED, consisting only of $7$ different objects, each of a different kind.
The selected objects are highlighted in red in Fig.~\ref{fig:icub-ds}.

We implement the classification algorithm using three different multiclass SVM versions, all based on a one-versus-all scheme: a linear SVM (denoted with \SVM in the following); an SVM with the RBF kernel (\SVMRBF); 
and an SVM with the RBF kernel implementing our defense mechanism based on rejection of adversarial examples (\SVMADV, Sect.~\ref{sec:improving_icub}).
The regularization parameter $C \in \{10^{-3}, \ldots, 10^{3}\}$ and the RBF kernel parameter $\gamma \in \{10^{-6}, \ldots, 10^{-2}\}$ have been set equal for all one-versus-all SVMs in each multiclass classifier, by maximizing recognition accuracy through 3-fold cross validation.

\begin{figure}[t]
\centering
\label{fig:no_attack_accuracy}
\includegraphics[width=0.99\columnwidth]{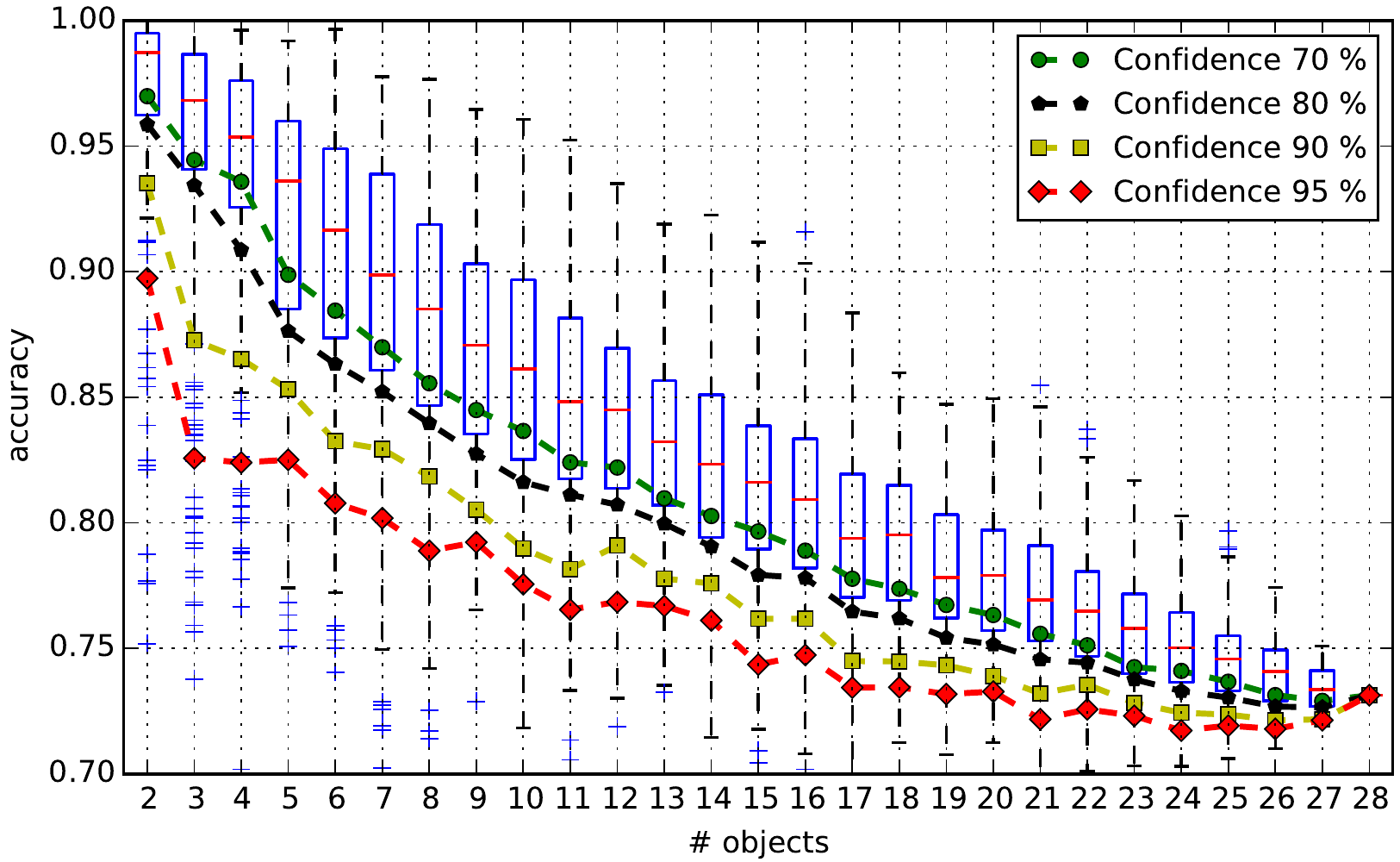}
\caption{Box plots of the recognition accuracies measured for linear SVM predictors trained on random subsets from $2$ to $28$ objects (whiskers with maximum 1.5 interquartile range). Dotted super-imposed curves represent the minimum accuracy guaranteed within a fixed confidence level.}
\label{fig:icub_base}
\end{figure}

\myparagraph{Baseline Performance.} In Fig.~\ref{fig:icub_base} we report a box plot showing the empirical probability distributions of the accuracy achieved by the \SVM classifier on increasingly larger object identification tasks, as suggested in~\cite{pasquale2015teaching}. To this end, we randomly select 300 subsets of increasing size from the \DIcub dataset (day4 acquisitions),  and then train and test the classifier on each subset. The achieved accuracy is considered an observation for estimating the empirical distributions. The minimum accuracy value for which the fraction of observations in the estimated distribution was higher than a specific confidence threshold is indicated as a dotted line. Notably the reported performances for the linear SVM are almost identical to those reported in \cite{pasquale2015teaching}, where a different algorithm is used.
Similar performances (omitted for brevity) are obtained using \SVMRBF.

\begin{figure*}[t]
\centering
\includegraphics[width=0.31\textwidth]{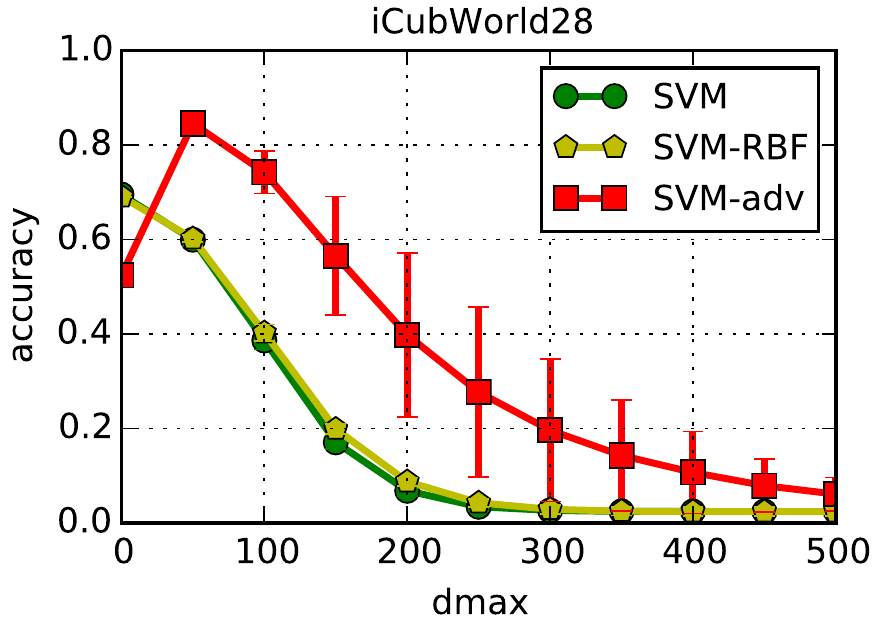}
\includegraphics[width=0.31\textwidth]{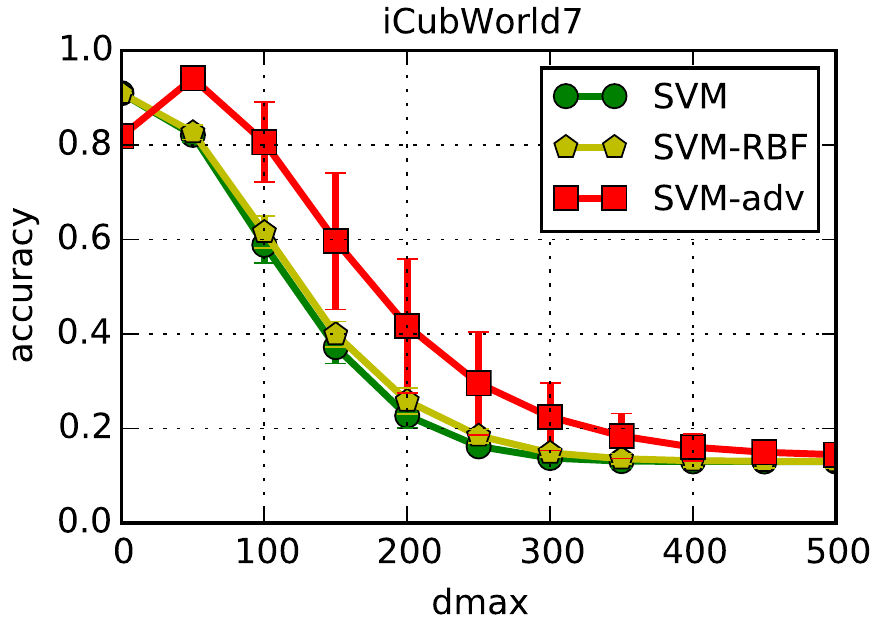}
\includegraphics[width=0.31\textwidth]{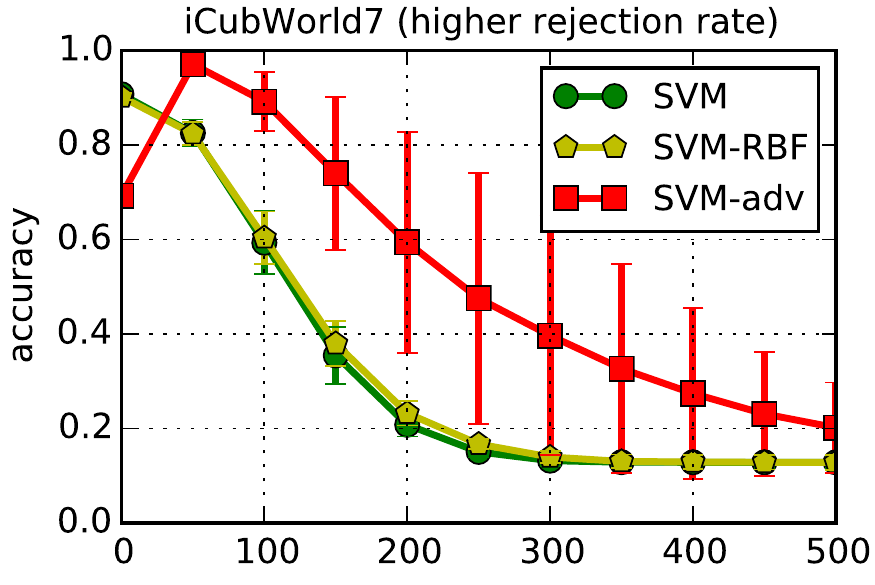}\\
\includegraphics[width=0.31\textwidth]{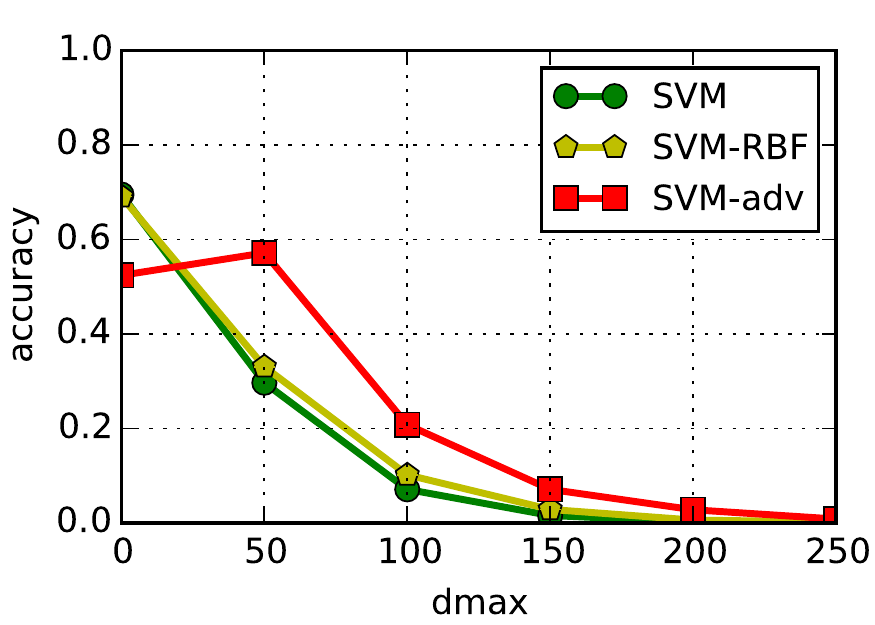}
\includegraphics[width=0.31\textwidth]{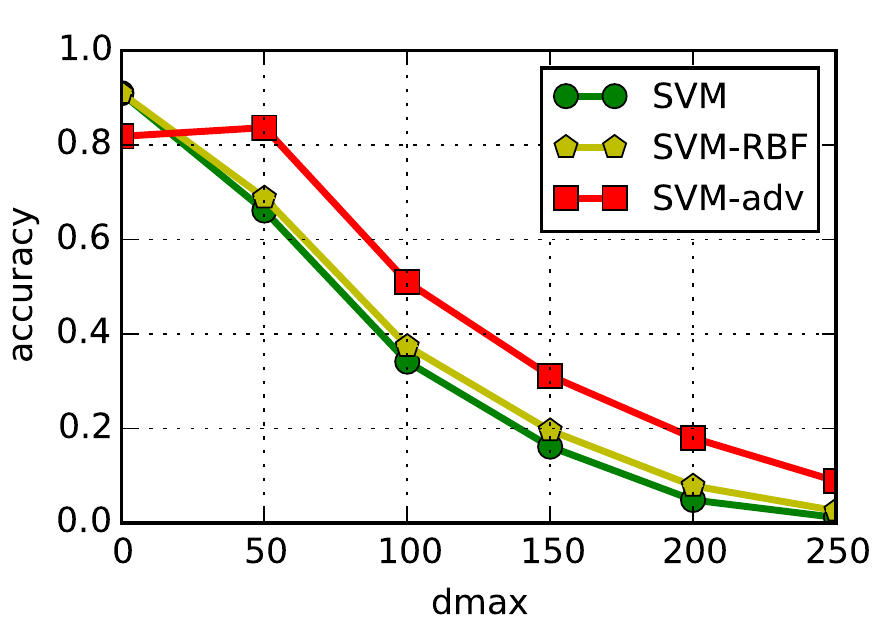}
\includegraphics[width=0.31\textwidth]{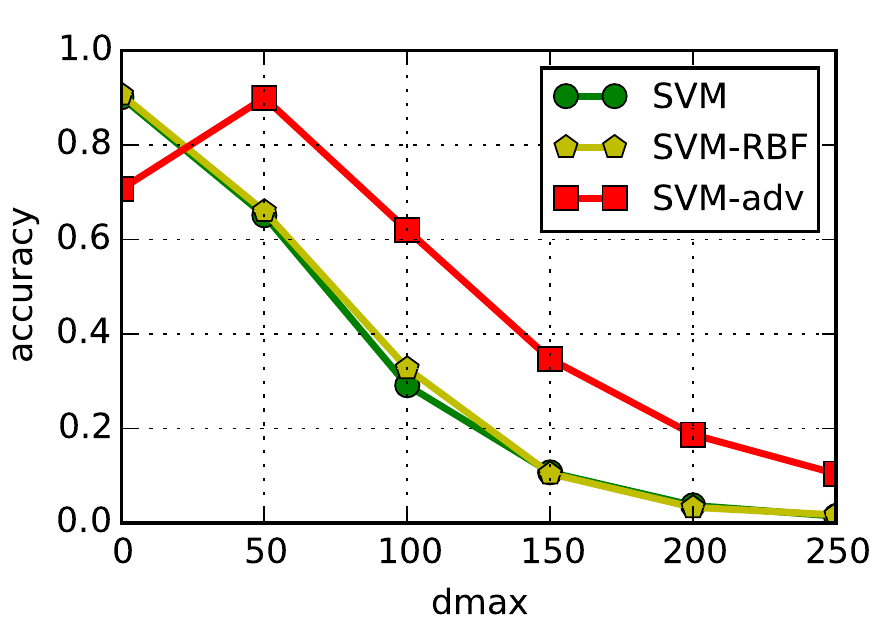}
\caption{Recognition accuracy of iCub (using the three different classifiers \SVM, \SVMRBF, and \SVMADV) against an increasing maximum admissible $\ell_{2}$ input perturbation $d_\text{max}$, for \DIcub (\emph{left column}) and \DIcubRED (\emph{middle} and \emph{right} columns), using error-specific (\emph{top row}), and error-generic (\emph{bottom row}) adversarial examples.
Baseline accuracies (in the absence of perturbation) are reported at $d_{\rm max}=0$.
}\label{fig:icub_seceval}
\end{figure*}

\myparagraph{Security Evaluation against Adversarial Examples.} We now investigate the security of iCub in the presence of adversarial examples. In this experiment, we consider the first 100 examples per class for both the \DIcub and \DIcubRED datasets, ending up with training and test sets consisting of $2,800$ and $700$ samples, respectively.
The recognition accuracy against an increasing maximum admissible $\ell_{2}$ perturbation (\ie, $d_{\rm max}$ value) is reported in Fig.~\ref{fig:icub_seceval} for both error-specific (top row plots) and error-generic (bottom row plots) attack scenarios. For error-specific evasion, we average our results not only on different training-test set splits, but also by considering a different target class in each repetition. While \SVM and \SVMRBF show a comparable decrease of accuracy at increasing $d_{\rm max}$, \SVMADV is able to strongly improve the security in most of the cases (as the corresponding curve decreases more gracefully). Notably, the performance of \SVMADV even increases for low values of $d_{\rm max}$. A plausible reason is that, even if all testing images are only slightly modified in input space, they immediately become blind-spot adversarial examples, ending up in a region which is far from the rest of the data.
As the input perturbation increases, such samples are gradually drifted inside a different class, becoming \emph{indistinguishable} from the samples of such class.

To further improve the security of iCub to adversarial examples, we set the rejection threshold of \SVMADV to a more conservative value, increasing the false negative rate for each base classifier of 5\% (estimated on a validation set). This results in a significant security improvement, as shown in the rightmost plots in Fig.~\ref{fig:icub_seceval}. However, as expected, this comes at the expense of misclassifying more legitimate (\ie non-manipulated) samples.

\myparagraph{Real-world Adversarial Examples.} In Fig.~\ref{fig:real_world_attacks} we report few adversarial examples generated using an error-specific evasion attack on the \DIcub data. Notably, the adversarial perturbation required to evade the system can be barely perceived by human eyes.
As an important real-world application of the proposed attack algorithm, in the bottom right plots of Fig.~\ref{fig:real_world_attacks}, we report an adversarial example generated by manipulating only a subset of the image pixels, corresponding to the label of the detergent. In this case, the perturbation becomes easier to spot for a human, but localizing the noise in a region of interest allows the attacker to construct a practical, real-world adversarial object, 
by simply attaching an ``adversarial'' sticker to the original object before showing it to the iCub humanoid robot.

\begin{figure*}[t]
\centering
\label{fig:real_world_attacks}
\includegraphics[width=0.22\textwidth]{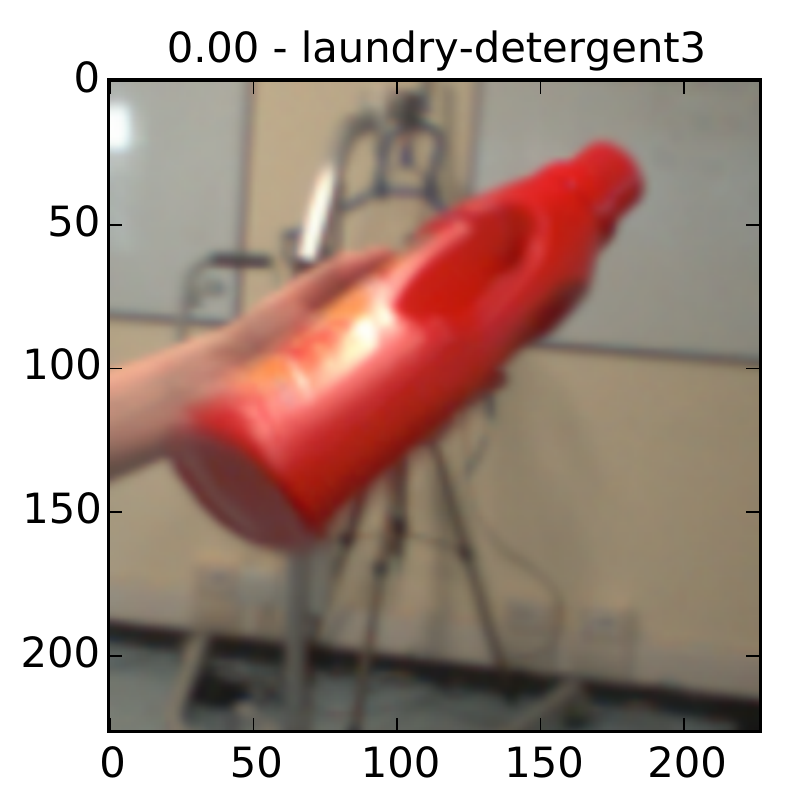}
\includegraphics[width=0.22\textwidth]{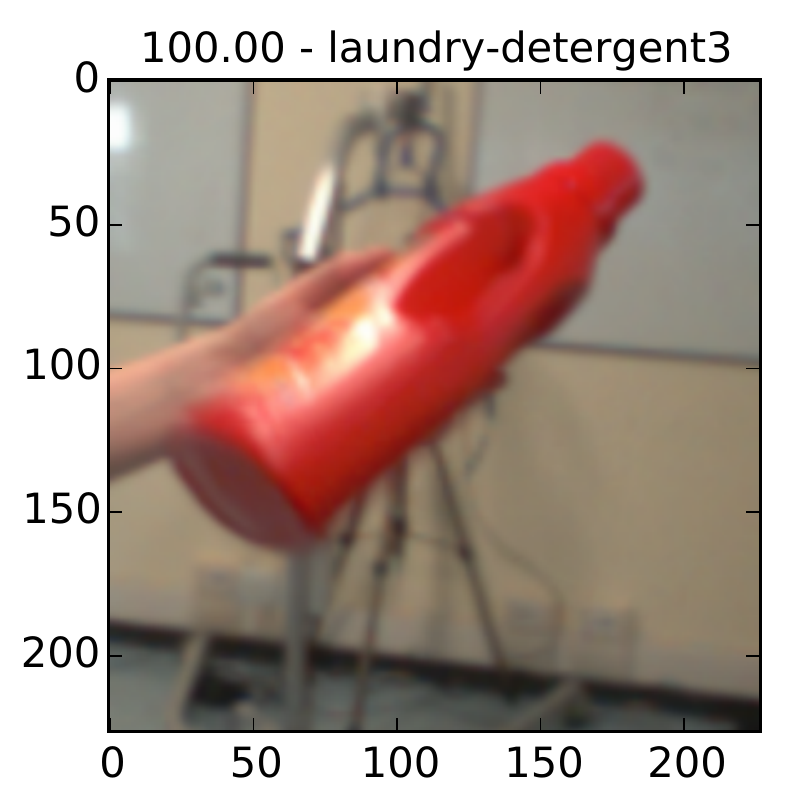}
\includegraphics[width=0.22\textwidth]{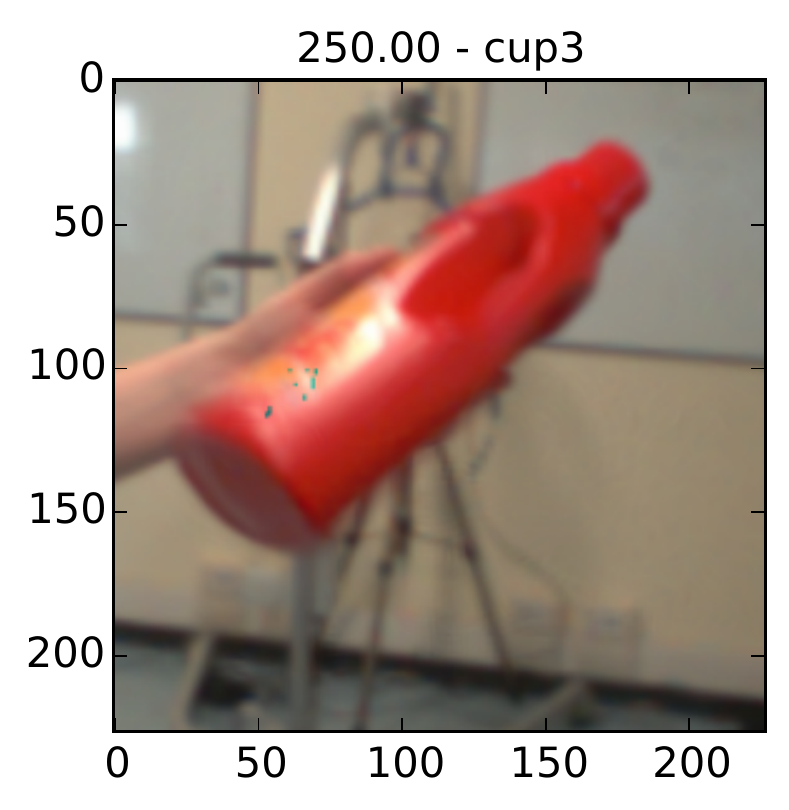}
\includegraphics[width=0.22\textwidth]{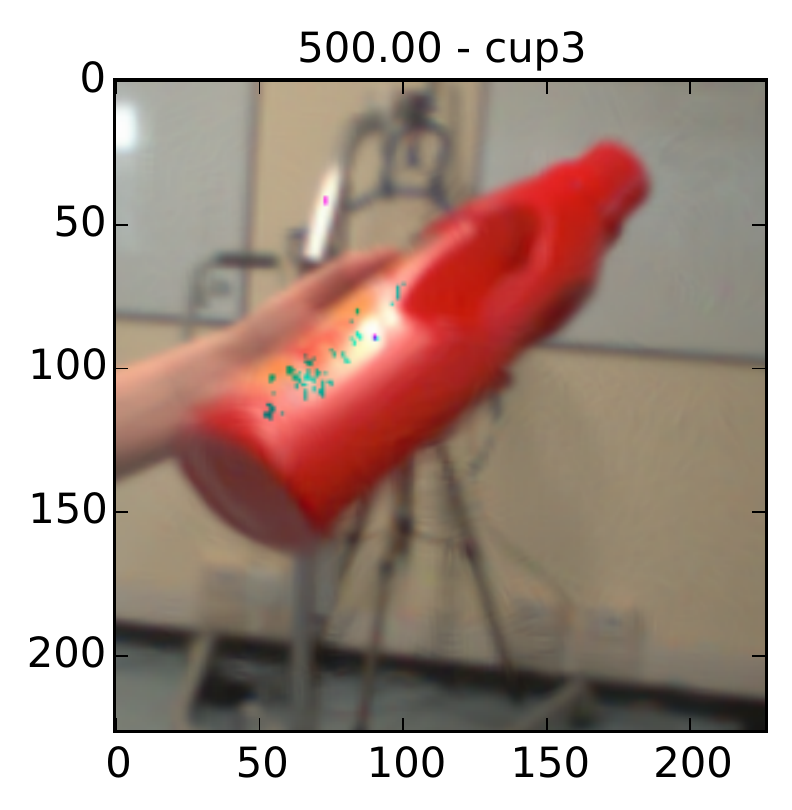}\\
\hspace{1pt}
\includegraphics[width=0.44\textwidth]{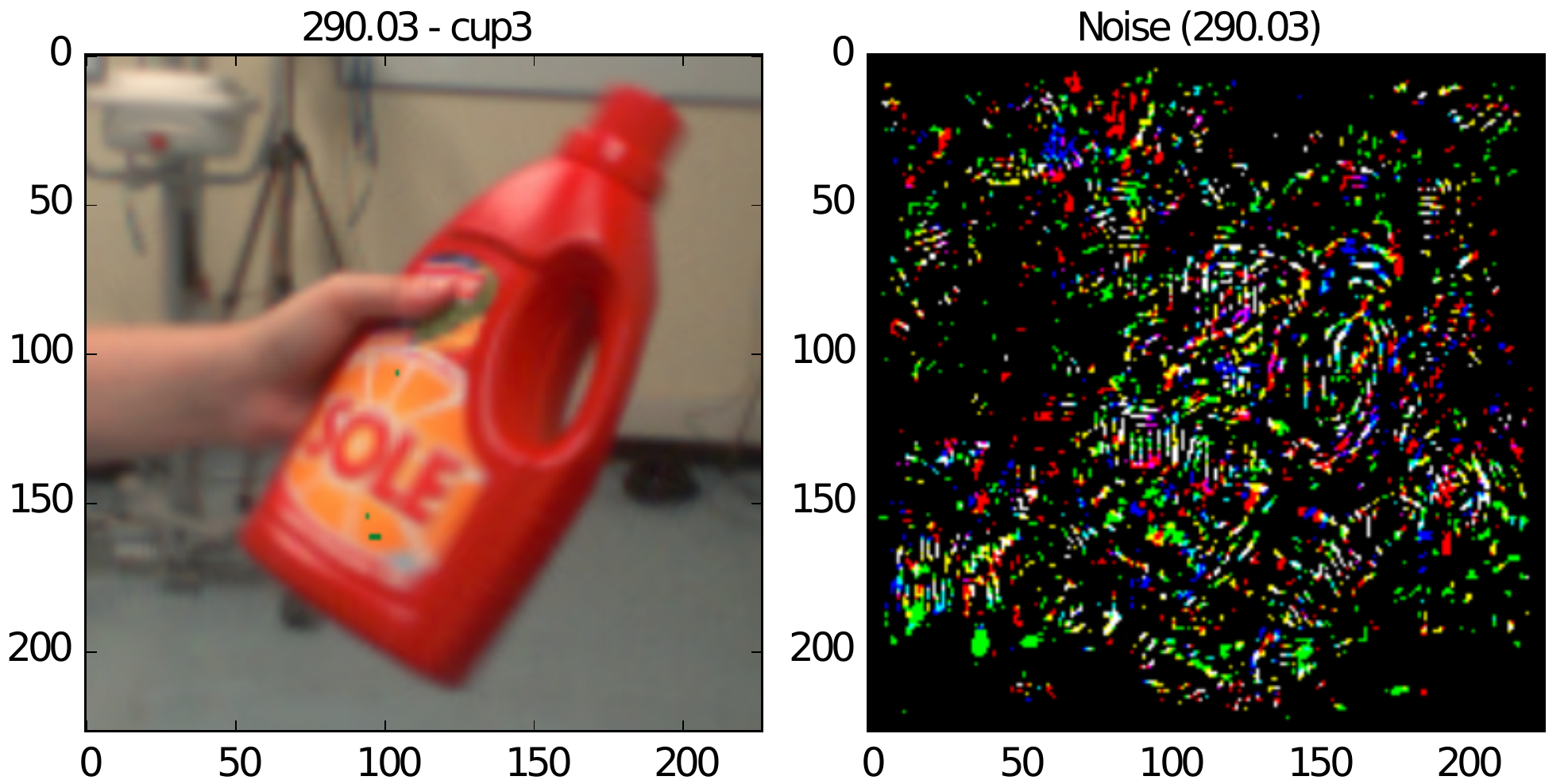}
\includegraphics[width=0.44\textwidth]{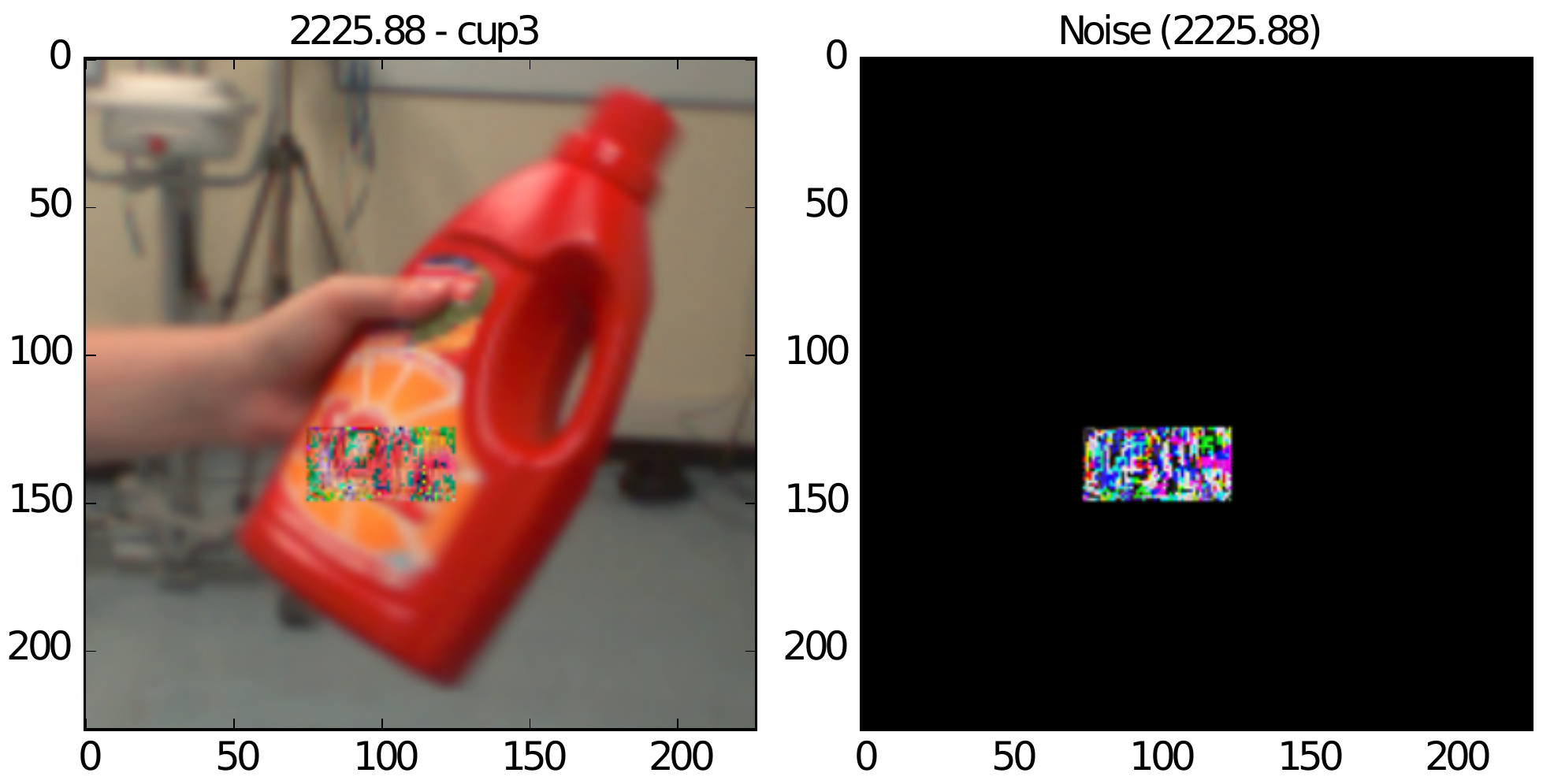}
\caption{
Plots in the top row show an adversarial example from class \emph{laundry-detergent3}, modified to be misclassified as \emph{cup3}, using an error-specific evasion attack, for increasing levels of input perturbation (reported in the title of the plots). 
Plots in the bottom row show the minimally-perturbed adversarial example that evades detection (\ie, the sample that evades detection with minimum $d_{\rm max}$), along with the corresponding noise applied to the initial image (amplified to make it clearly visible), for the case in which all pixels can be manipulated (first and second plot), and for the case in which modifications are constrained to the label of the detergent (\ie, simulating a sticker that can be applied to the real-world \emph{adversarial} object).}
\label{fig:real_world_attacks}
\end{figure*}

\myparagraph{Why are Deep Nets Fooled?} Our analysis shows that also the iCub vision system can be fooled by adversarial examples, even by only adding a slightly-noticeable noise to the input image.
To better understand the root causes of this phenomenon, we now provide an empirical analysis of the sensitivity of the feature mapping induced by the ImageNet deep network used by iCub, by comparing the $\ell_{2}$ distance corresponding to random and adversarial perturbations in the input space, with the one measured in the deep feature space. To this end, we randomly perturb each training image such that the $\ell_{2}$ distance between the initial and the perturbed image in the input space equals $10$. We then measure the  $\ell_{2}$ distance between the deep feature vectors corresponding to the same images. For randomly-perturbed images, the average distance in deep space (along with its standard deviation) is $0.022 \pm 0.002$, while for the adversarially-perturbed images, it is $2.386 \pm 0.386$.
This means that random perturbations in the input space only result in a very small shift in the deep space, while even light alterations of an image along the adversarial direction cause a large shift in deep space, which in turn highlights a significant \emph{instability} of the deep feature space mapping induced by the ImageNet network.
In other words, this means that images in the input space are very close to the decision boundary along the adversarial (gradient) direction, as conceptually represented in Fig.~\ref{fig:dnn_sensitivity}. Note that this is a general issue for deep networks, not only specific to ImageNet~\cite{tanay16-arxiv,feinman17-arxiv,szegedy14-iclr, goodfellow15-iclr,moosavi2016deepfool,papernot16-sp}.

\begin{figure*}[t]
	\label{fig:dnn_sensitivity}
	\centering
	\includegraphics[width=0.23\textwidth]{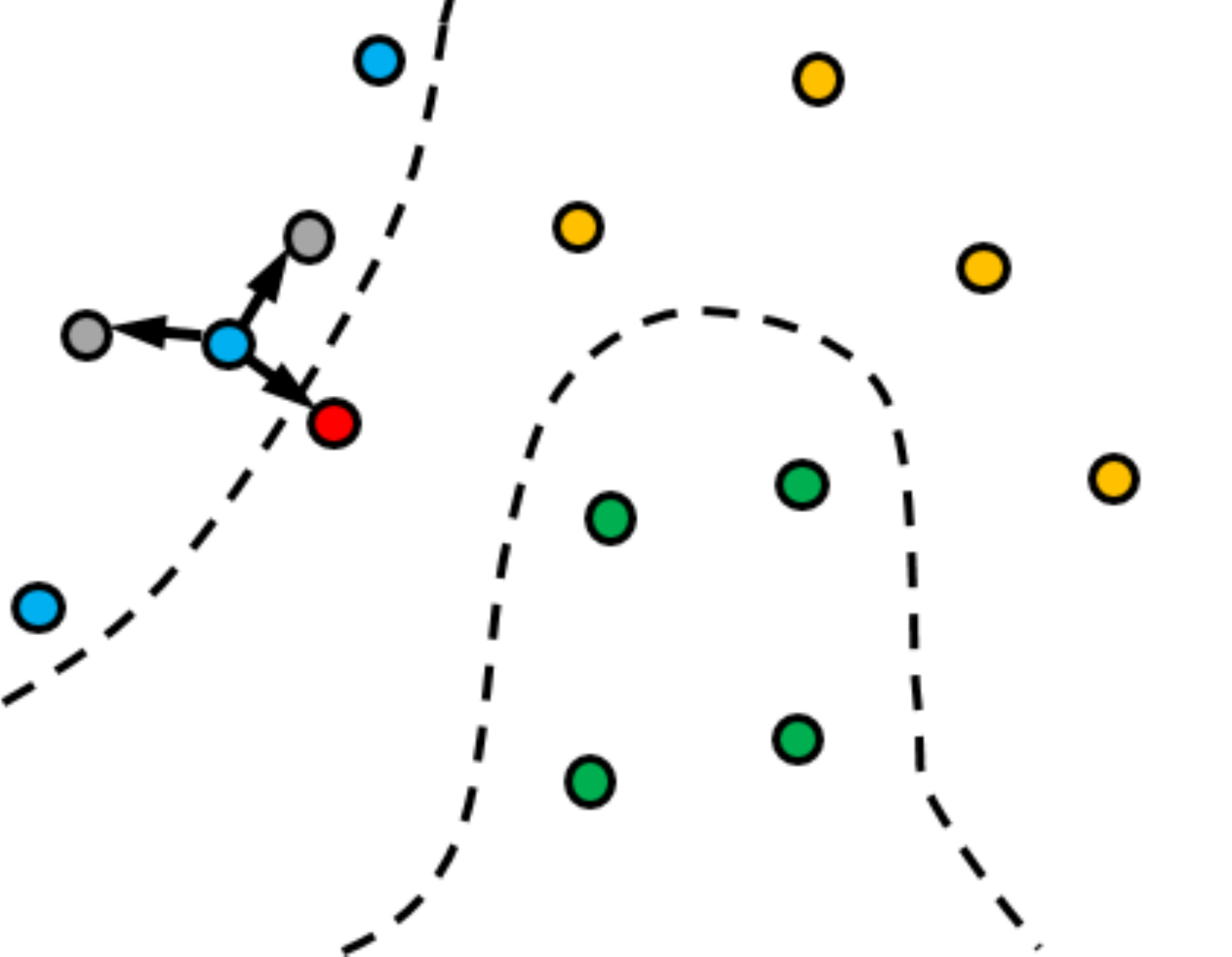}
	\hspace{20pt}
	\includegraphics[width=0.23\textwidth]{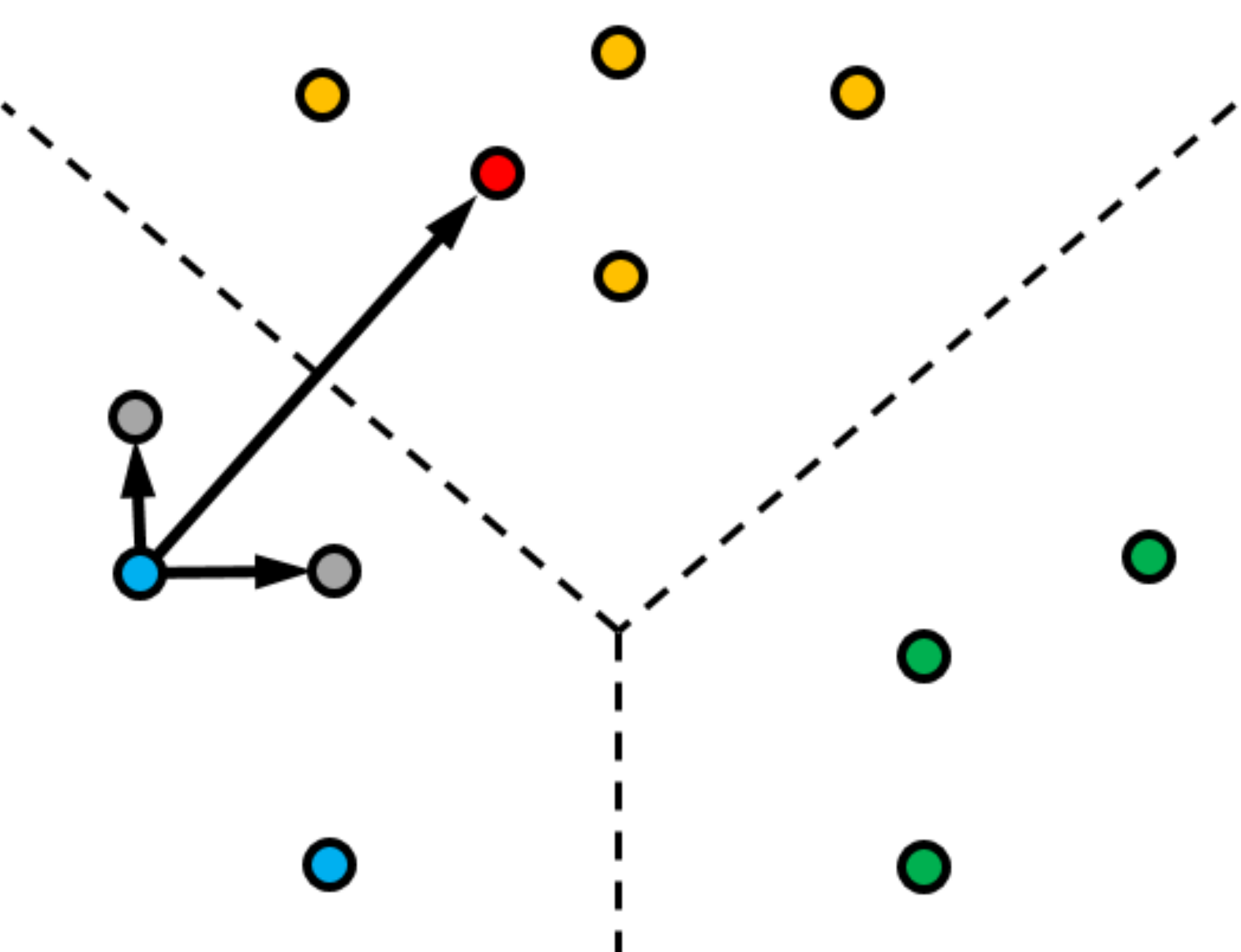}
	\caption{
	Conceptual representation of the vulnerability of the deep feature space mapping. The left and right plots respectively represent images in input space and the corresponding deep feature vectors.
	Randomly-perturbed versions of an input image are shown as gray points, while the adversarially-perturbed image is shown as a red point. Despite these points are at the same distance from the input image in the input space, the adversarial example is much farther in the deep feature space. This also means that images are very close to the decision boundary in input space, although in an adversarial direction that is difficult to guess at random due to the high dimensionality of the input space.}\label{fig:dnn_sensitivity}
\end{figure*}

It should be thus clear that even a well-crafted modification of the last layers of the network, as in our proposed defense mechanism \SVMADV, can only mitigate this vulnerability. Indeed, it remains intimately related to the stability of the deep feature space mapping, which can be only addressed by imposing specific constraints while training the deep neural network; \eg, by imposing that small shifts in the input space correspond to small changes in the deep space, as recently proposed in~\cite{zheng2016improving}. Another possible countermeasure to improve stability of such mapping is to enforce classification of samples within a minimum \emph{margin}, by modifying the neurons' activation functions and, potentially, considering a different regularizer for the objective function optimized by the deep network. In this respect, it would be interesting to investigate more in detail the intimate connections between robustness to adversarial input noise and regularization, as highlighted in~\cite{xu09,russu16-aisec}.

\section{Related Work} 
\label{sec:related_work}
Previous work has investigated the problem of adversarial examples in deep networks~\cite{szegedy14-iclr, goodfellow15-iclr,moosavi2016deepfool,papernot16-sp,tanay16-arxiv,feinman17-arxiv}, focusing however on minimally-perturbed adversarial examples, \ie, examples that simply lie inside the decision region of a known class, even if they remain far from the corresponding training examples; on the contrary, our approach is based on creating maximally-perturbed (indistinguishable) adversarial examples (misclassified with high confidence).
Different techniques aimed at improving the security of deep networks have also been proposed.
Some of them attempt to reduce classifier vulnerability by directly detecting and rejecting adversarial examples \cite{bendale2016towards,Li16-cvpr}.
The technique of \cite{bendale2016towards} is based on open-set recognition: it rejects samples whose distance from the centroids of all the classes exceeds a given threshold. However, it has not been evaluated using adversarial examples carefully generated to evade the classifier.
In \cite{Li16-cvpr}, adversarial examples are detected using the output of the first convolutional layers. A different approach has been proposed in \cite{zheng2016improving}: it aims at improving the stability of the deep feature space mapping by retraining the network using an objective function that penalizes examples (images) that are close in input space but lie far in deep feature space. This approach has however been investigated only against small image distortions.

\section{Conclusions and Future Work}
\label{sec:conclusion}

Deep learning has shown groundbreaking performance in several real-world application domains, encompassing areas like computer vision, speech recognition and language processing, among others.
Despite its impressive performances, recent work has shown how deep neural networks can be fooled by well-crafted adversarial examples affected by a barely-perceivable adversarial noise.
In this work, we have developed a novel algorithm for the generation of adversarial examples which enables a more complete evaluation of the security of a learning algorithm, and apply it to investigate the security of the robot-vision system of the iCub humanoid. 
Even if we do not restrict ourselves to the manipulation of pixels belonging to the object of interest in the image (which could lead one to more easily generate the corresponding real-world adversarial object, \eg, by mean of the application of specific stickers to objects), we have shown how our algorithm enables this additional possibility.
Notably, even if we have not constructed any real-world adversarial object during our experiments, recent work has shown that the artifacts introduced by printing images and re-acquiring them through a camera are irrelevant, and do not eliminate the problem of the existence of adversarial examples~\cite{kurakin2016adversarial}.
Similarly, another work has shown how to evade face recognition systems based on deep learning by using adversarial glasses and other accessories~\cite{sharif2016accessorize}. These recent evidences clearly give a much higher practical relevance to the problem of adversarial examples.

We have demonstrated and quantified the vulnerability of iCub to the presence of adversarial manipulations of the input images, and suggested a simple countermeasure to mitigate the threat posed by such an issue. We have additionally shown that, while blind-spot adversarial examples  can be detected using our defense mechanism, to further improve the security of iCub against \emph{indistinguishable} adversarial examples, re-training the classification algorithm on top of a pre-trained deep neural network is not sufficient.
To this end, different strategies to enforce the deep network to learn a more stable deep feature representation (in which small perturbations to the input data correspond to small perturbations in the deep feature space) should also be adopted, like the one proposed in~\cite{zheng2016improving}.

Other interesting research directions for this work include evaluating security of robot-vision systems against other threats, including the threat of data poisoning~\cite{huang11,biggio12-icml,biggio14-tkde}, in which a malicious human annotator may provide few wrong labels to the humanoid to completely mislead its learning process and enforce it to misclassify as many objects as possible.
In general, a comprehensive, standardized framework for evaluating the security of such systems while providing also more formal verification procedures is still lacking, and we believe that this constitutes a fundamental requirement for the complete transition of deep-learning-based systems in safety-critical applications, like robots performing life-critical tasks and self-driving cars.



\end{document}